\documentclass[conference]{IEEEtran}
\usepackage{times}

\usepackage[numbers]{natbib}
\usepackage{multicol}
\usepackage[bookmarks=true]{hyperref}

\usepackage{helvet}
\usepackage{courier}
\usepackage{algorithm}
\usepackage{algorithmic}
\usepackage{multirow}
\usepackage{graphics}
\usepackage{amsmath}
\usepackage{amsfonts}

\newtheorem{definition}{Definition}

\DeclareMathOperator*{\argmin}{\arg\!\min}

\begin{document}

\title{Plan Explicability and Predictability for \\Robot Task Planning}

% You will get a Paper-ID when submitting a pdf file to the conference system
\author{Yu Zhang, Sarath Sreedharan, Anagha Kulkarni, Tathagata Chakraborti, \\ Hankz Hankui Zhuo and Subbarao Kambhampati}

%\author{\authorblockN{Michael Shell}
%\authorblockA{School of Electrical and\\Computer Engineering\\
%Georgia Institute of Technology\\
%Atlanta, Georgia 30332--0250\\
%Email: mshell@ece.gatech.edu}
%\and
%\authorblockA{Twentieth Century Fox\\
%\and
%\authorblockN{James Kirk\\ and Montgomery Scott}
%\authorblockA{Starfleet Academy\\
%San Francisco, California 96678-2391\\

% avoiding spaces at the end of the author lines is not a problem with
% conference papers because we don't use \thanks or \IEEEmembership

%Homer Simpson\authorrefmark{2},
%Montgomery Scott\authorrefmark{3} and
%Eldon Tyrell\authorrefmark{4}}
%Telephone: (800) 555--1212, Fax: (888) 555--1212}
%\authorblockA{\authorrefmark{4}Tyrell Inc., 123 Replicant Street, Los Angeles, California 90210--4321}}

\maketitle

%%%%%%%%%%%%%%%%%%%%%%%%%%%%%%%%%%%%%%%%%%%%%%%%%%%%%%%
%%%%%%%%%%%%%%%%%%%%%%%%%%%%%%%%%%%%%%%%%%%%%%%%%%%%%%%
\begin{abstract}
Intelligent robots and machines are becoming pervasive in human populated environments.
A desirable capability of these agents is to respond to
goal-oriented commands by autonomously constructing task plans. 
However, such autonomy can add significant cognitive load 
and potentially introduce safety risks to humans when agents 
behave unexpectedly. 
Hence, for such agents to be helpful,  
one important requirement is for them to synthesize plans that can be easily understood by humans.
While there exists previous work that studied socially acceptable robots
that interact with humans in ``natural ways'',
and work that investigated legible motion planning,  
there lacks a general solution for high level task planning.
To address this issue, we introduce the notions of plan {\it explicability} and {\it predictability}. 
To compute these measures, 
first, we postulate that humans understand agent plans by associating abstract tasks with agent actions,
which can be considered as a labeling process. 
We learn the labeling scheme of humans for agent plans from training examples
using conditional random fields (CRFs). 
Then, we use the learned model to label a new plan to compute its explicability and predictability.
These measures can be used by agents to proactively choose
or directly synthesize plans
that are more explicable and predictable to humans.
We provide evaluations on a synthetic domain and 
with human subjects using physical robots to show the effectiveness of our approach.
\end{abstract}

\IEEEpeerreviewmaketitle

%%%%%%%%%%%%%%%%%%%%%%%%%%%%%%%%%%%%%%%%%%%%%%%%%%%%%%%
%%%%%%%%%%%%%%%%%%%%%%%%%%%%%%%%%%%%%%%%%%%%%%%%%%%%%%%
\section{Introduction}

Intelligent robots and machines are becoming pervasive in human populated environments. 
Examples include robots for education, entertainment and personal assistance just to name a few. 
Significant research efforts have been invested to build autonomous agents to make them more helpful. 
These agents respond to 
goal specifications instead of basic motor commands,
which requires them to autonomously synthesize
task plans and execute those plans to achieve the goals. 
However, if the behaviors of these agents are incomprehensible,
it can increase the cognitive load of humans and 
potentially introduce safety risks to them. 

As a result,
one important requirement for such intelligent agents is to ensure
that the synthesized plans are comprehensible to humans. 
This means that instead of considering only the 
planning model of the agent,
%(e.g., which allows agents to synthesize a plan to achieve a given goal), 
plan synthesis should also consider the 
interpretation 
of the agent behavior from the human's perspective.
This interpretation is related to our modeling of other agents. 
More specifically, we tend to have expectations of others' behaviors based on
our understanding (modeling) of their capabilities, mental states and etc. 
If their behaviors do not match with these expectations,
we would often be confused. 
One of the major reasons of this confusion is due to the fact
that our understanding of others' models is often partial and inaccurate.
This is also true when humans interact with intelligent agents.
%as well as the human's preferences for the agent behaviors. 
For example, to darken a room that is too bright,
a robot can either adjust the window blinds, 
switch off the lights,
or break the light bulbs in the room.
While breaking the light bulbs may well be the least costly plan to the robot under certain conditions
(e.g., when the robot cannot easily move in the environment but we are unaware of it), 
it is clear that the other two options are far more desirable in the context of robots cohabiting with humans. 
One of the challenges here is that 
the human's understanding of the agent model
%how the human interprets the agent's model
%(which determines how the robot chooses its plan in the above example) 
is inherently hidden.
Thus, its interpretation from the human's perspective can be arbitrarily different from the agent's own model. 
While there exists previous work that studied socially acceptable robots \cite{aas,aasflu,Shah,hri}
that interact with humans in ``natural ways'',
and work that investigated legible motion planning \cite{Dragan-RSS-13},
there lacks a general solution for high level task planning.
%especially for robots and intelligent agents that can construct their own plans.

In this paper, we introduce the notions of plan {\it explicability}
and {\it predictability}
which are used by autonomous agents (e.g., robots) to synthesize ``explicable plans'' 
that can be easily understood by humans. 
%Plan explicability is related to the trust issue in automation. 
%It is well known that automation can have both positive and negative effects on human performance.
%Many human factor studies have shown that automation should be designed to benefit human-machine interaction \cite{visser-cedm-2011};
%it is argued in \cite{christoffersen-hpce-2002} that automation must be provided in the context of interaction.
%These prior works provide the motivation for generating explicable plans in this work.
Our problem settings are as follows:
an intelligent agent is given a goal by a human (so that the human knows the goal of the agent) working in the same environment
and it needs to synthesize a plan to achieve the goal.
As suggested in psychological studies \cite{Vallacher,csibra2007obsessed},
we assume that humans naturally interpret a plan as achieving abstract tasks (or subgoals),
which are functional interpretations of agent action sequences in the plan.
%Given that most humans only have a limited understanding of agent model,
For example, a robot that executes a sequence of manipulation actions
may be interpreted as achieving the task of ``$picking$ $up$ $cup$''.
%There are studies \cite{} supporting this kind of interpretations. 
Based on this assumption, intuitively, %given any prefix of a plan (as a sub-plan), 
the easier it is for humans to associate tasks with actions in a plan,
the more explicable the plan is.
 %sub-plan;
Similarly, the easier it is to predict the next task given actions in the previous tasks,
the more predictable the plan is.
In this regard, explicability is concerned with the association between human-interpreted tasks and agent actions,
while predictability is concerned with the connections between these abstract tasks. 
%One observation is that plan predictability is partially relying on plan explicability.
%Intuitively, it is difficult to predict the next task of the agent if 
%we do not even understand what it is currently trying to achieve.

Since the association between tasks and agent actions can be considered as a labeling process, 
we learn the labeling scheme of humans for agent plans
from training examples using conditional random fields (CRFs).
We then use the learned model to label a new plan to compute its explicability and predictability.
These measures are used by agents to proactively choose  
or directly synthesize plans 
that are more explicable and predictable without affecting the quality much. 
Our learning approach does not assume any prior knowledge on the human's interpretation of the agent model.
%The learned CRF can keep evolving through human feedbacks.
We provide evaluation on a synthetic domain in simulation and with human subjects using physical robots to demonstrate the effectiveness of our approach.

%%%%%%%%%%%%%%%%%%%%%%%%%%%%%%%%%%%%%%%%%%%%%%%%%%%%%%%
%%%%%%%%%%%%%%%%%%%%%%%%%%%%%%%%%%%%%%%%%%%%%%%%%%%%%%%
\section{Related Work}

\begin{figure}
\centering
{
    \includegraphics{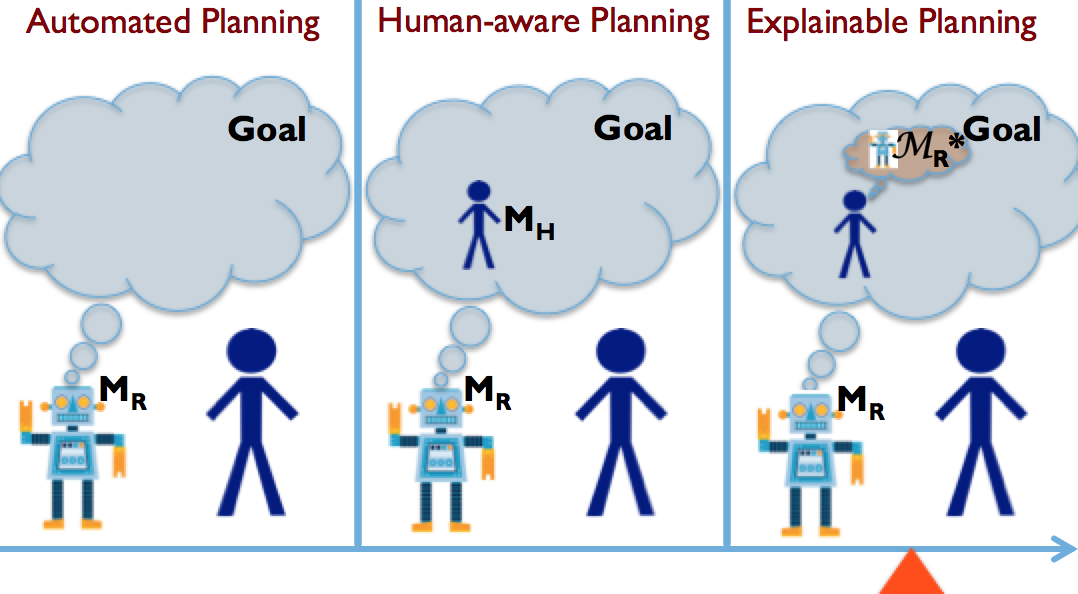}
}
\caption{
%General problem settings. 
%The left part shows that an agent plan based solely on $M_R$ has parts that are inexplicable and unpredictable (in terms of topological distance) from the human's perspective (i.e., in $\mathcal{M}_R^*$). 
%The middle part shows a plan that is more explicable but still not very predictable.
%The right part shows a plan that is both explicable and predictable. 
%The human's interpretation ($\mathcal{M}_R^*$) is assumed to be associated with tasks (i.e., $T_1 - T_3$).
%The agent model ($M_R$) is assumed to be specified as transitions between states. 
%The relationships between tasks and states are shown at the bottom. 
%The human interprets the agent goal $G$ as achieving the three tasks $T_1 - T_3$.
From left to right, the scenarios illustrate the differences between automated task planning, human-aware planning
and explicable planning (this work). 
In human-aware planning, the robot needs to maintain a model of the human (i.e., $M_H$) which captures the human's capabilities, intents and etc.
In explicable planning, the robot considers the differences between its model from the human's perspective (i.e., $\mathcal{M}_R^*$)
and its own model $M_R$.
}
\label{fig:exp}
\end{figure}

To build autonomous agents (e.g., robots),
one desirable capability is for such agents to respond to goal-oriented commands via automated task planning. 
A planning capability allows agents to autonomously synthesize
plans to achieve a goal given the agent model ($M_R$ as shown in the first scenario in Fig. \ref{fig:exp}) 
instead of following low level motion commands,
thus significantly reducing the human's cognitive load. 
Furthermore, to work alongside of humans, 
these agents must be ``human-aware'' when synthesizing plans.
In prior works, this issue is addressed under human-aware planning \cite{sisbot-tro-2007,cirillo,serendipity}
in which agents take the human's activities and intents into account when constructing their plans. 
This corresponds to human modeling in human-aware planning
as shown in the second scenario in Fig. \ref{fig:exp}.
A prerequisite for human-aware planning is a plan recognition component,
which is used to infer the human's goals and plans.
This information is then used to avoid interference, 
and plan for serendipity and teaming with humans.
There exists a rich literature on plan recognition \cite{489919,Charniak199353,ramirez2010plan,icapsnew},
and many recent works use these techniques 
in human-aware planning and human-robot teaming \cite{kartik-iros-2014,serendipity,yz-IROS15}.

While our work on plan explicability and predictability 
falls within the scope
of human-in-the-loop planning (which also includes human-aware planning), 
it differs significantly from the previous work. 
This is illustrated in Fig. \ref{fig:exp}. 
More specifically, in human-aware planning, 
the challenge is to obtain the human model ($M_H$ in Fig. \ref{fig:exp})
which captures human capabilities \cite{capmodel}, intents \cite{kartik-iros-2014,serendipity} and etc. 
%it is assumed that the agent ($R$) has a model of the human ($H$).  
%While this model ($M_H$ in Fig. \ref{fig:exp}) is arguably obtainable 
%via prior knowledge, communication or learning when assuming a specific model representation \cite{hankz-noisy},
%it is difficult to argue the same in our context. 
The modeling in this work is one level deeper:
it is about the interpretation of the agent model from the human's perspective ($\mathcal{M}_R^*$ in Fig. \ref{fig:exp}).
In other words,
$R$ needs to understand the model of {\it itself} in $H$'s eyes.
This information is inherently hidden, difficult to convey,
and can be arbitrarily different (e.g., having different representations) from $R$'s own model ($M_R$ in Fig. \ref{fig:exp}). 

There exists work on generating legible robot motions \cite{Dragan-RSS-13}
which considers a similar issue in motion planning. 
We are, on the other hand, concerned with task planning. 
Note that two different task plans may map to exactly the same motions which can be interpreted vastly differently by humans.
In such cases, considering only motion becomes insufficient. 
Nevertheless, there exists similarities between \cite{Dragan-RSS-13} and our work.
For example, legibility there is analogous to predictability in ours.

%Note first that plan explicability is a fundamental issue regardless of whether the human knows about the goal of the agent or not.
%Meanwhile, plan explicability is related to goal and plan recognition --
%if a human does not know what an agent is trying to achieve (i.e., the goal of the agent), 
%the agent behavior is more likely to be inexplicable. 
%To clearly distinguish our work from goal recognition, in our context, we assume that the human 
%knows the goal of the agent.

In the human-robot interaction (HRI) community,
there exists prior works that discuss how to enable natural and fluent human-robot interaction
\cite{aas,aasflu,Shah,hri} to create 
more socially acceptable robots \cite{Fong_2003_4157}.
These works, however, apply only to
behaviors in specific domains.
Compared with model learning via expert teaching, 
such as inverse reinforcement learning \cite{Abbeel} and tutoring systems \cite{murray},
which is about learning the ``right'' model from teachers,
%However, there are no experts (or teachers) involved in our context.
%More specifically, although the intelligent agent knows more than the human at the low level,
%it needs to consider the human's high level preferences. 
our work, on the other hand, is concerned with learning model differences.
%Our work is also connected to preferred grammar in HTN planning \cite{Li}
%since model differences ultimately lead to plan preference.  
Furthermore, as an extension to our work, 
when robots cannot find an explicable plan that is also cost efficient,
they need to explain the situation. 
In this regard, our work is also related to excuse \cite{ICAPS101453} and explanation generation \cite{Hanheide2015}.
Finally, while our learning approach appears to be similar to information extraction \cite{Peng2006963},
we use the learned model to proactively guide planning
instead of passively extracting information.

%%%%%%%%%%%%%%%%%%%%%%%%%%%%%%%%%%%%%%%%%%%%%%%%%%%%%%%
%----------------------------------------------------------------------------------------------------------------------------------------------
\section{Explicability and Predictability}

In our settings, an agent $R$ needs to achieve a goal given by a human
in the same environment (so that the human knows about the goal of the robot). 
The agent has a model of itself (referred to as $M_R$)
which is used to autonomously construct plans to achieve the goal.
In this paper, we assume that this model is based on PDDL \cite{fox-jair-2003},
a general planning domain definition language. 
As we discussed, for an agent to generate explicable and predictable plans,
it must not only consider $M_R$
but also $\mathcal{M}_R^*$, 
which is the interpretation of $M_R$ from the human's perspective.
%Given this background, we next discuss the problem formulation. 
%a general view of the problem and 
%then discuss our solution. 

%----------------------------------------------------------------------------------------------------------------------------------------------
\subsection{Problem Formulation}

Keeping the problem settings in mind, 
%The general problem settings are shown in Fig. \ref{fig:exp},
%are represented graphically. 
%$\mathcal{M}_R^*$ is presented as thought clouds and $M_R$ is presented in the polygonal area. 
given a domain, the problem is to find a plan for a given goal that satisfies the following:
\begin{equation}
\argmin_{\pi_{M_R}}{cost(\pi_{M_R}) + \alpha \cdot dist(\pi_{M_R}, \pi_{\mathcal{M}_R^*})}
\label{equ:opt}
\end{equation}
where $\pi_{M_R}$ is a plan that is constructed using $M_R$ (i.e., the agent's plan),
$\pi_{\mathcal{M}_R^*}$ is a plan that is constructed using $\mathcal{M}_R^*$ (i.e., the human's anticipation of the agent's plan),
$cost$ returns the cost of a plan,
$dist$ returns the distance (i.e., capturing the differences) between two plans,
and $\alpha$ is the relative weight.
The goal of Eq. \eqref{equ:opt} is to find a plan that minimizes a weighted sum of 
the cost of the agent plan and the differences between the two plans. 
Since the agent model $M_R$ is assumed to be given, 
the challenge lies in the second part in Eq. \eqref{equ:opt}.

Note that if we know $\mathcal{M}_R^*$ or it can be learned,
the only thing left would be to search for a proper $dist$ function.
%we can solve Eq. \eqref{equ:opt}.
However, as discussed previously, $\mathcal{M}_R^*$ is inherently hidden,
difficult to convey,
and can be arbitrarily different from $M_R$.
Hence, our solution is to use a learning method to directly approximate the returned values. 
We postulate that humans understand agent plans by associating abstract tasks with actions,
which can be considered as a labeling process.
Based on this, we assume that $dist(\pi_{M_R}, \pi_{\mathcal{M}_R^*})$ can be functionally decomposed as:
\begin{equation}
dist(\pi_{M_R}, \pi_{\mathcal{M}_R^*}) = F \circ {\mathcal{L}}^*(\pi_{M_R})
\label{equ:decomp}
\end{equation}
where $F$ is a domain specific function that takes plan labels as input,
and $\mathcal{L}^*$ is the labeling scheme of the human for agent plans based on $\mathcal{M}_R^*$.
As a result, Eq. \eqref{equ:opt} now becomes:
\begin{equation}
\argmin_{\pi_{M_R}}{cost(\pi_{M_R}) + \alpha \cdot F \circ \mathcal{L}^*_{CRF}(\pi_{M_R} | \{S_i|S_i = {\mathcal{L}}^*(\pi^i_{M_R})\})}
\label{equ:opt-crf}
\end{equation}
where $\{S_i\}$ is the set of training examples
and $\mathcal{L}^*_{CRF}$ is the learned model of $\mathcal{L}^*$.
We can now formally define plan explicability and predictability in our context.
Given a plan of agent $R$ as a sequence of actions, 
we denote it as $\pi_{M_R}$ and simplified below as $\pi$ for clarity:
\begin{equation}
	\pi = \langle a_0, a_1, a_2,... a_N\rangle
\end{equation}
where $a_0$ is a null action that denotes plan starting.
Given the domain, 
we assume that a set of task labels $T$ is provided to label agent actions:
\begin{equation}
	T = \{T_1, T_2,... T_M\}
\end{equation}

\subsubsection{Explicability Labeling}
Explicability is concerned with the association between abstract tasks and agent actions;
each action in a plan is associated with an action label.
The set of action labels for explicability is the power set of the task labels:
%\footnote{Note that to provide more flexibility in task labeling for humans
%we can incorporate a natural language understanding method to cluster texts and associate them with the underlying tasks in arbitrary domains.}:
\begin{equation}
	L = 2^T
\label{equ:label}
\end{equation}
When an action label includes multiple task labels, the action is interpreted as contributing to multiple tasks;
when an action label is the empty set, the action is interpreted as inexplicable.
When a plan is labeled, 
we can compute its 
explicability measure based on its action labels in a domain specific way. 
More specifically, we define:
\begin{definition}[Plan explicability]
Given a domain, the explicability $\theta_\pi$ of an agent plan $\pi$ is computed by
	a mapping, $F_{\theta}: {\textbf{L}}_\pi \rightarrow [0, 1]$ (with $1$ being the most explicable).
\label{def:ex}
\end{definition}
${\textbf{L}}_\pi$ above denotes the sequence of action labels for $\pi$.
An example of $F_\theta$ used in our evaluation is given below:
\begin{equation}
F_\theta({\textbf{L}_\pi}) = \frac{\sum_{i \in [1, N]}{\textbf{1}}_{L(a_i)\neq{\emptyset}}}{N}
\label{def:eva-ex}
\end{equation}
where $N$ is the plan length, $L(a_i)$ returns the action label of $a_i$, 
and ${\textbf{1}}_{formula}$ is an indicator function that returns $1$ when the $formula$ holds or $0$ otherwise.
Eq. \eqref{def:eva-ex} basically computes the ratio between the number of actions with non-empty action labels
and the number of all actions.

\subsubsection{Predictability Labeling}
Predictability is concerned with the connections between tasks in a plan. 
An action label for predictability is composed of two parts:
a current label and a next label (i.e., $L \times L$).
The current label is also the action label for explicability.
The next label (similar to the current label) is used to specify the tasks that are anticipated to be achieved next.
A next label with multiple task labels is interpreted as having multiple candidate tasks to achieve next;
when this label is the empty set,  it is interpreted as that the next task is unpredictable,
or there are no more tasks to be achieved.

\begin{definition}[Plan Predictability]
Given a domain, the predictability $\beta_\pi$ of a plan $\pi$ is computed by
a mapping, $F_\beta: {\textbf{L}}^2_\pi \rightarrow [0, 1]$ (with $1$ being the most predictable).
\label{def:pd}
\end{definition}
${\textbf{L}}^2_\pi$ denotes the sequence of action labels for predictability. 
An example of $F_\beta$ is given below which is used in our evaluation when assuming that the current and next labels are associated with at most one task label:
\begin{equation}
F_\beta({\textbf{L}^2_\pi}) = \frac{\sum_{i \in [0, N]}{\textbf{1}}_{|L(a_i)| = 1} \land ({\textbf{1}}_{L^2(a_i)=L(a_j)} \lor {\textbf{1}}_{L^2(a_{i:N})={\emptyset}})}{N + 1}
\label{def:eva-pd}
\end{equation}
where 
$a_j (j > i)$ is the first action that has a different current label as $a_i$
or the last action in the plan if no such action is unfound,
$L^2(a_i)$ returns the next label of $a_i$
and ${\textbf{1}}_{L^2(a_{i:N})={\emptyset}}$ returns $1$ only if the next labels for all actions after $a_i$ (including $a_i$) are $\emptyset$.
Eq. \eqref{def:eva-pd} computes the ratio between number of actions that we have correctly predicted the next task
and the number of all actions. 

%----------------------------------------------------------------------------------------------------------------------------------------------
\subsection{A Concrete Example}

\begin{figure}
	\centering
	{
		\includegraphics{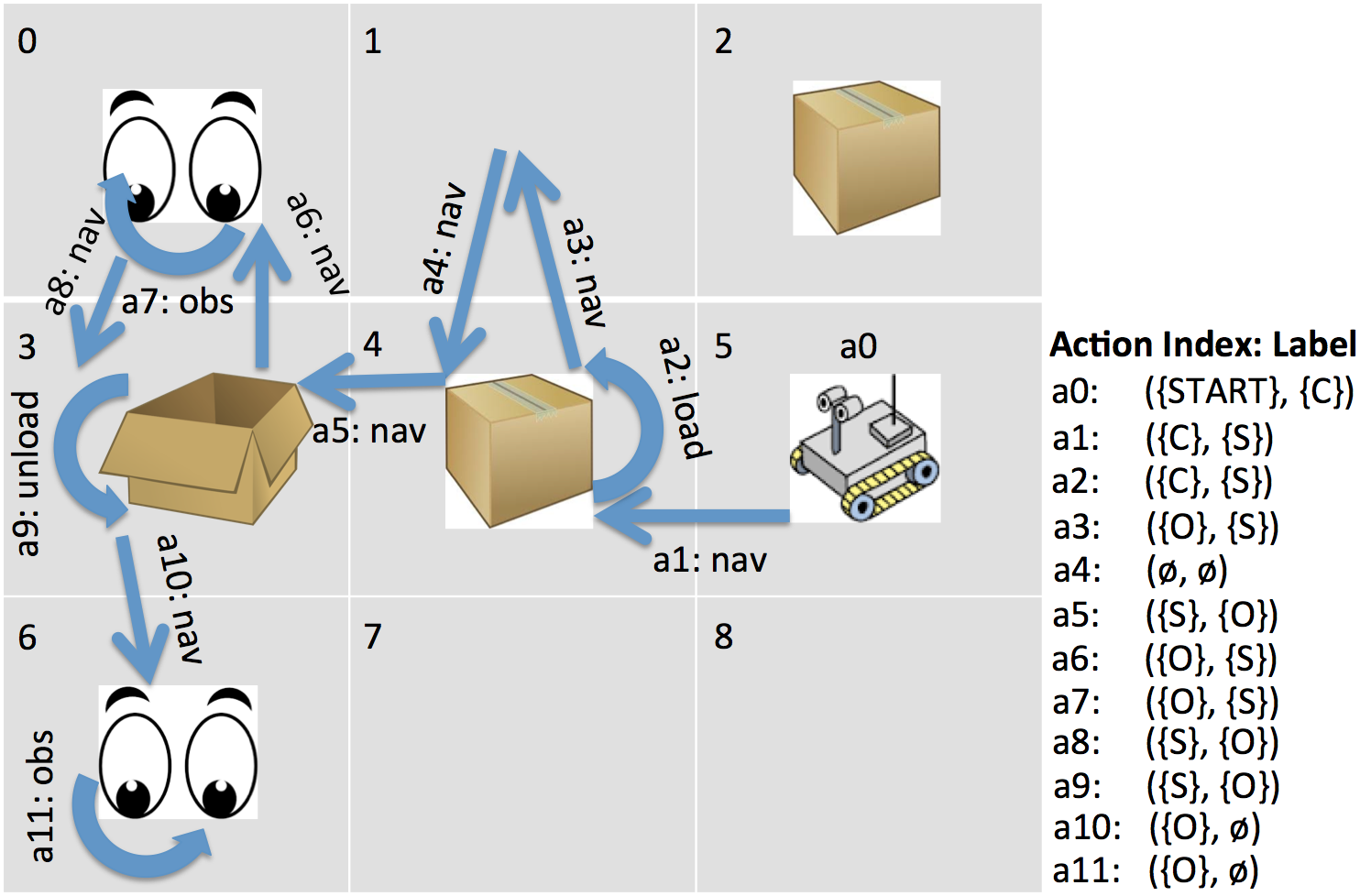}
	}
	\caption{Example for plan explicability and predictability with action labels (on the right) for a given plan in the rover domain.}
	\label{fig:example}
\end{figure}

Before discussing how to learn the labeling scheme of the human from training examples,
we provide a concrete example to connect the previous concepts 
and show how training examples can be obtained.
In this example, there is a rover in a grid environment working with a human.
An illustration of this example is presented in Fig. \ref{fig:example}.
There are resources to be collected
which are represented as boxes.
There is one storage area that can store one resource
which is represented as an open box. 
The rover can also make observations.
The rover actions include $\{navigate$ $l_{from}$ $l_{to}\}$, $\{observe$ $l\}$
$\{load$ $l\}$, and $\{unload$ $l\}$,
each representing a set of actions since $l$ (i.e., representing a location) can be instantiated to different locations (i.e., $0-8$ in Fig. \ref{fig:example}).
%Each $l$ $(l \in \{0-8\})$ with or without a subscript denotes a location. 
$navigate$ (or $nav$) can move the rover from a location
to one of its adjacent locations;
$load$ can be used to pick up a resource when the rover is not already loaded; 
$unload$ can be used to unload a resource at a storage area if the area is empty; 
$observe$ (or $obs$) can be used to make an observation.
Once a location is observed, 
it remains observed. 
The goal in this example is for the rover to make 
the storage area non-empty and observe two locations
that contain the eye symbol in Fig. \ref{fig:example}.

In this domain, we assume that there are three
abstract tasks that may be used by the human to interpret the rover's plans:
COLLECT (C), STORE (S) and OBSERVE (O).
Note that we do not specify any arguments 
for these tasks (e.g., which resource the rover is collecting)
since this information may not be important to the human.
This also illustrates that $M_R$ and $\mathcal{M}_R^*$ can be arbitrarily different.
In Fig. \ref{fig:example},
we present a plan of the rover as connected arrows
starting from the its initial location.

{\it Human Interpretation as Training Examples:}
Let us now discuss how humans may 
interpret this plan (i.e., associating labels with actions)
as the actions are observed {\it incrementally}:
when labeling $a_i$, we only have access to the plan prefix $\langle a_0, ..., a_i\rangle$.
At the beginning for labeling $a_0$, the observation is that the rover starts at $l_5$.
Given the environment and knowledge of the rover's goal, 
we may infer that the first task should be COLLECT 
(the resource from $l_4$). 
Hence, we may choose to label $a_0$ as ({\{START\}, \{C\})}.
The first action of the rover (i.e., $nav$ $l_5$ $l_4$)
seems to match with our prediction. 
Furthermore, given that the storage area is closest to the rover's location after completing COLLECT, 
the next task is likely to be STORE.
Hence, we may label $a_1$ as (\{C\}, \{S\}) as shown in the figure.
The second action (i.e., $load$ $l_4$) also matches with our expectation.
Hence, we label $a_2$ too as (\{C\}, \{S\}).
The third action,  
$nav$ $l_4$ $l_1$, 
however, is unexpected since we predicted STORE in the previous steps.
Nevertheless, we can still explain it as contributing to OBSERVE (at location $l_0$). 
Hence, we may label this navigation action ($a_3$) as (\{O\}, \{S\}). 
For the fourth action, the rover
moves back to $l_4$, which is inexplicable since the rover's behavior seems to be oscillating without particular reasons.
Hence, we may choose to label this action as ($\emptyset$, $\emptyset$).
The labeling for the rest of the plan continues in a similar manner.
This thought process reflects how training examples can be obtained from human labelers. 
%When humans are to label a plan in robotic applications,
%they will also be given the actual plan execution by the robot (e.g., as video streams).

%----------------------------------------------------------------------------------------------------------------------------------------------
\section{Learning Approach}

To compute $\theta_\pi$ and $\beta_\pi$ from Defs. \eqref{def:ex} and \eqref{def:pd} for a given plan $\pi$, 
the challenge is to provide a label for each action.
This requires us to learn the labeling scheme of humans (i.e., $\mathcal{L}^*$ in Eq. \eqref{equ:decomp}) from training examples 
and then apply the learned model to $\pi$ (i.e., $\mathcal{L}_{CRF}^*$ in Eq. \eqref{equ:opt-crf}).
To formulate a learning method, we consider the sequence of labels as hidden variables.
The plan that is executed by the agent (which also captures the state trajectory), 
as well as any cognitive cues that may be obtained (e.g., from sensing) during the plan execution 
constitute the observations.
The graphical model that we choose for our learning approach is conditional random fields (CRFs) \cite{Lafferty}
due to their abilities to model sequential data. 
An alternative would be HMMs; however, CRFs have been shown to relax assumptions about the input and output sequence distributions
 and hence are more flexible. 
 
The distributions that are captured by CRFs have the following form:
\begin{equation}
	p(\textbf{x}, \textbf{y}) = \frac{1}{Z}\Pi_{A}\Phi(\textbf{x}_A, \textbf{y}_A)
\end{equation}
in which $Z$ is a normalization factor that satisfies:
\begin{equation}
Z = \sum_{\textbf{x}, \textbf{y}}\Pi_{A}\Phi(\textbf{x}_A, \textbf{y}_A)
\end{equation}

In the equations above, $\textbf{x}$ represents the sequence of observations,
$\textbf{y}$ represents the sequence of hidden variables,
and $\Phi(\textbf{x}_A, \textbf{y}_A)$ represents a factor 
that is related to a subgraph in the CRF model associated with variables $\textbf{x}_A$ and $\textbf{y}_A$.
In our context, $\textbf{x}$ are the observations made during the execution of a plan; 
$\textbf{y}$ are the action labels. 
Each factor is associated with a set of features that can be extracted during the plan execution.
Next, we discuss some possible features that can be used for plan explicability and predictability.

%----------------------------------------------------------------------------------------------------------------------------------------------
\subsection{Features for Learning}

Given an agent plan, the immediate set of features that we have access to is 
the plan and its associated state trajectory. 
Note that the human may not be required (nor it is necessary) to fully understand this information.
%as long as it captures what the human can interpret.
%these features should capture changes that humans can interpret.
When the dynamics of the agent are known,
given the plan,
it may also be possible to derive low level motor commands that implement the motions,
which can be used to extract motion related features.

When the agent is equipped with sensors such as cameras and lasers, 
we can also extract features from sensor information.
For example, from video streams and depth information,
we can extract features about the environment, 
e.g., how crowded the workspace is.

Sensor information can also be used to extract dynamic features such as the location of the human. 
However, note that this information will not be available during the testing phase,
%During plan synthesis (i.e., testing phase), 
%since our goal is to generate explicable and predictable plans proactively before they are executed,
%sensor information will be missing. 
and thus these features need to be estimated based on other information 
(e.g., projected plan of the human based on plan recognition techniques \cite{ramirez2010plan,icapsnew}).

In this work, we use a linear chain CRF.
However, our formulation is easily extensible to more general types of CRFs.
Given an agent plan $\pi = \langle a_0, a_1, a_2, ...\rangle$, 
each action is associated with a set of features. 
Hence, each training example is of the following form: 
\begin{equation}
	\langle (F_0, L^2_0), (F_1, L^2_1), (F_2, L^2_2), ... \rangle
\label{eq:training_example}
\end{equation}
where $L^2_i$ is the action label for predictability (and explicability) for $a_i$. $F_i$ is the set of features for $a_i$.
We discuss several feature categories in more detail below:

\subsubsection{Plan Features}
Given the agent model (specified in PDDL),
the set of plan features for $a_{i}$ includes the action description
and the state variables 
after executing the sequence of actions $\langle a_0, ..., a_{i}\rangle$
from the initial state.
This information can be easily extracted given the model.
For example, in our rover example in Fig. \ref{fig:example},
this set of features for $a_1$ includes
$navigate$, $at$ $rover$ $l_4$, $at$ $resource0$ $l_2$, $at$ $resource1$ $l_4$, $at$ $storage0$ $l_3$.

\subsubsection{Action Features}
Action features
for $a_i$ describes the motion (e.g., dynamics) of this action.
These features can be used to capture, for example, 
smoothness of execution within and across actions.
Action features sometimes serve as important cognitive cues for humans to 
understand agent actions.  
For example, an action that enables a robot to cross a river may be 
interpreted as {\it swimming}, {\it pedaling}, or {\it propelling}
depending on how the robot motion looks like.
Action features can be extracted for a plan
given the dynamics of the robot.

\subsubsection{Interaction Features}
Interaction features are intended to capture $a_i$'s influence
on the human.
For example, it can include how far the agent is from human
and what the human is performing when $a_i$ is being executed.
In other words, this set of features captures characteristics of the interactions between the human and agent.
%and how these interactions influence the human's interpretation of the agent's plan
%under different situations.
Interaction features can be extracted from sensor information
or estimated based on the projected human plan.
 
%During plan synthesis or the testing phase when such information is unavailable, 
%this set of features may need to be estimated,
%for example, based on the 
%inferred goal and plan of the human. 

%----------------------------------------------------------------------------------------------------------------------------------------------
\subsection{Using the Learned Model}

Given a set of training examples in the form of Eq. \eqref{eq:training_example},
we can train the CRF model to learn the labeling scheme in Eq. \eqref{equ:opt-crf}.
%Given a plan, this model is used to predict the labels
%which are then used to compute its $\theta$ and $\beta$ measures. 
We discuss two ways to use the learned CRF model. 

\subsubsection{Plan Selection}
The most straightforward method is to perform plan selection on a set of candidate plans
which can simply be a set of plans that are within a certain cost bound of the optimal plan. 
Candidate plans can also be generated to be diverse with respect to various plan distances. 
For each plan, 
the agent must first extract the features of the actions as we discussed earlier. 
It then uses the trained model (i.e., $\mathcal{L}_{CRF}^*$) to produce the labels for the actions in the plan. 
$\theta$ and $\beta$ can then be computed given the mappings in Defs. \eqref{def:ex} and \eqref{def:pd}.
These measures can then be used to choose a plan that is more explicable and predictable.

\subsubsection{Plan Synthesis}
A more efficient way is to incorporate these measures as heuristics into the planning process.
Here, we consider the FastForward (FF) planner with enforced hill climbing \cite{Hoffmann:2001}. 
To compute the heuristic value given a planning state, 
we use the relaxed planning graph to construct the remaining planning steps. 
However, since relaxed planning does not ensure a valid plan,
%the planning steps beyond the current planning state may not
%necessarily lead to valid states.
we can only use action descriptions as plan features for actions
that are beyond the current planning state 
when estimating the $\theta$ and $\beta$ measures.
These estimates are then combined with the relaxed planning heuristic 
(which only considers plan cost)
to guide the search.
The algorithm for generating explicable and predictable plans 
is presented in Alg \ref{alg}.
%The algorithms for generating predictable plans
%and for considering both explicability and predictability measures
%can be similarly constructed.

\begin{algorithm}[!ht]
\caption{Synthesizing Explicable and Predictable Plans}
\textbf{Input:} agent model $M_R$, trained human labeling scheme $\mathcal{L}_{CRF}^*$, initial state $I$ and goal state $G$.\\
\textbf{Output:} $\pi_{EXP}$ 
\begin{algorithmic}[1]
\STATE Push $I$ into the open set $O$.
\WHILE{open set is not empty} 
	\STATE $s = $ GetNext($O$).
	\STATE $h^* = MAX$.
	\IF{$G$ is reached}
		\RETURN $s$.plan (i.e., the plan that leads to $s$ from $I$).
	\ENDIF 
	\STATE Compute all possible next states $N$ from $s$.
	\FOR{$n \in N$}
		\STATE Compute the relaxed plan $\pi_{RELAX}$ for $n$.
		\STATE Concatenate $s$.plan (with plan features) with $\pi_{RELAX}$ (with only action descriptions) as $\bar{\pi}$.
		\STATE Compute and add other relevant features. 
		\STATE Compute $\textbf{L}^2_\pi = \mathcal{L}_{CRF}^*(\bar{\pi})$.
		\STATE Compute $\theta$ and $\beta$ based on $\textbf{L}^2_\pi$ for $\bar{\pi}$.
		\STATE Compute $h = f(\theta, \beta, h_{cost})$ ($f$ is a combination function; $h_{cost}$ is the relaxed planning heuristic).
	\ENDFOR
	\STATE Find the state $n^* \in N$ with the minimum $h$.
	\IF{$h(n^*) < h^*$} 
		\STATE Clear $O$.
		\STATE Push $n^*$ into $O$.
	\ELSE
		\STATE Push all $n \in N$ into $O$.
	\ENDIF
\ENDWHILE
\end{algorithmic}
\label{alg}
\end{algorithm}
 
The capability to synthesize explicable and predictable plans is useful for autonomous agents.
For example, in domains where humans interact closely with robots (e.g., in an assembly warehouse),
more preferences should be given to plans that are more explicable and predictable
since there would be high risks if the robots act unexpectedly. 
One note is that the relative weights of explicability and predictability may vary in different domains.  
For example, in domains where robots do not engage in close interactions with humans,
predictability may not matter much.
%We plan to investigate the integration of these measures in various domains.

%%%%%%%%%%%%%%%%%%%%%%%%%%%%%%%%%%%%%%%%%%%%%%%%%%%%%%%
%----------------------------------------------------------------------------------------------------------------------------------------------
\section{Evaluation}

We first evaluate our approach systematically
on a synthetic dataset
based on the rover domain. 
Then, we evaluate it with human subjects using physical robots to validate that
the synthesized plans are more explicable to humans in a blocks world domain. 
%the learned model can indeed capture patterns of ``explicability'' in a car driving domain.

\subsection{Systematic Evaluation with a Synthetic Domain}

The aim is twofold here: evaluate how well the learning approach can capture an arbitrary labeling scheme;
evaluate the effectiveness of plan selection and synthesis with respect to the $\theta$ and $\beta$ measures. 

\subsubsection{Dataset Synthesis}

To simplify the data synthesis process, we make the following assumptions: 
all rover actions have the same cost;
all rover actions are associated with at most one task label (i.e., $L = T \cup \{\emptyset\}$ in Eq. \eqref{equ:label}). 
To construct a domain in which the optimal plan (in terms of cost)
may not be the most explicable (in order to differ $M_R$ from $\mathcal{M}_R^*$), 
we add ``oscillations'' to the plans of the rover.
These oscillations are incorporated by randomly adding 
locations for the rover to visit as {\it hidden goals}.
For these locations, the rover only needs to visit them.
As a result, it may demonstrate ``unexpected'' behaviors 
given only the {\it public goal}, denoted by $G$, which is known to both the rover and human.
We denote the goal that also includes the hidden goals as $G'$.
Given a problem with a public goal $G$, 
we implement a labeling scheme to automatically provide the ``ground truth'' of a rover plan,
which is constructed by the rover to achieve $G'$.

Given a plan of the rover, we label it incrementally by associating each action with a current and next label.
These labels are chosen from
\{\{COLLECT\}, \{STORE\}, \{OBSERVE\}, $\emptyset$\}.
We denote the plan prefix $\langle a_0, ... a_i \rangle$ for a plan $\pi$ as $\pi_{i}$,
the state after applying $\pi_{i}$ as $s_i$ from the initial state,
and a plan that is constructed from $s_i$ to achieve $G$ 
(i.e., using $s_i$ as the initial state) as $P(s_i)$.
For the current label of $a_i$:
\begin{enumerate}
	\item If $|P(s_i)| \geq |P(s_{i-1})|$, we label $a_i$ as $\emptyset$ (i.e., inexplicable).
	This rule means that humans may label an action as inexplicable
	if it does not contribute to achieving $G$. 
	\item If $|P(s_i)| < |P(s_{i-1})|$, we label $a_i$ based on the distances from the current rover location to the targets 
	(i.e., storage areas or observation locations),
	current state of the rover (i.e., loaded or not), and whether $a_i$ moves the rover closer to these targets. 
	For example, if the closest target is a storage area and the rover is loaded,
	we label $a_i$ as \{STORE\}. 
	When there are ties, 
	we label $a_i$ as $\emptyset$ (i.e., unclear and hence interpreted as inexplicable).
\end{enumerate}
For the next label of $a_i$:
\begin{enumerate}
	\item  
	This label is determined by the target that is closest to the rover state after the current task is achieved.
	When there are ties, $a_i$ is labeled as $\emptyset$ (i.e., unclear and hence interpreted as unpredictable).
	If the current label is $\emptyset$, 
	we also label $a_i$ as $\emptyset$ (i.e., unpredictable). 
	\item If the current task is also the last task, we label $a_i$ as $\emptyset$ since there is no next task.
\end{enumerate}
For evaluation, we define $F_\theta$ and $F_\beta$ as in Eqs. \eqref{def:eva-ex} and \eqref{def:eva-pd}.
%\begin{equation}
%F_\theta({\textbf{L}_\pi}) = \frac{\sum_{i \in [1, N]}{\textbf{1}}_{L(a_i)\neq{\emptyset}}}{N}
%\label{def:eva-ex}
%\end{equation}
%where $N$ is the plan length, $L(a_i)$ returns the action label of $a_i$, 
%and ${\textbf{1}}_{formula}$ is an indicator function that returns $1$ when the $formula$ holds or $0$ otherwise.
%\begin{equation}
%F_\beta({\textbf{L}^2_\pi}) = \frac{\sum_{i \in [0, N]}{\textbf{1}}_{|L(a_i)| = 1} \land ({\textbf{1}}_{L^2(a_i)=L(a_j)} \lor {\textbf{1}}_{L^2(a_{i:N})={\emptyset}})}{N + 1}
%\label{def:eva-pd}
%\end{equation}
%where 
%$a_j (j > i)$ is the first action that has a different current label as $a_i$
%or the last action in the plan if unfound,
%$L^2(a_i)$ returns the next label of $a_i$
%and ${\textbf{1}}_{L^2(a_{i:N})={\emptyset}}$ returns $1$ only if the next labels for all actions after $a_i$ (including $a_i$) are $\emptyset$ . 
We randomly generate problems in a $4 \times 4$ environment. 
For each problem, we randomly generate $1-3$ resources as a set RE,
$1-3$ storage areas as a set ST,
$1-3$ observation locations as a set OB.
The public goal $G$ of a problem, first, includes making all storage areas non-empty.
To ensure a solution, we force $|RE| = |ST|$ if $|RE| < |ST|$.
Furthermore, the rover must make observations at the locations in OB. 
$G'$ for the rover includes $G$ above, as well as a set of hidden goals. 
Locations of the rover, RE, ST, OB and hidden goals are randomly generated in the environment  
and do not overlap in the initial state.
Although seemingly simple, the state space of this domain is on the order of $10^{20}$.

\subsubsection{Results}

%Since we use synthetic data,
%we extract only plan features in our evaluations.
We use only plan features here. 
%First, we evaluate the predication performance of the learned model (i.e., $\mathcal{L}_{CRF}^*$) as the number of training samples increases.
First, we evaluate our approach to learning the labeling scheme (i.e., $\mathcal{L}_{CRF}^*$) 
as the difference between $M_R$ and $\mathcal{M}_R^*$ 
gradually increases (i.e., as the number of hidden goals increases). 
Afterwards, we evaluate the effectiveness of plan selection and synthesis with respect to the $\theta$ and $\beta$ measures. 
To verify that our approach can generalize to different problem settings, 
we fix the level of oscillation when generating training samples while allowing it to vary in testing samples. 

\begin{figure}
\centering
{
    \includegraphics{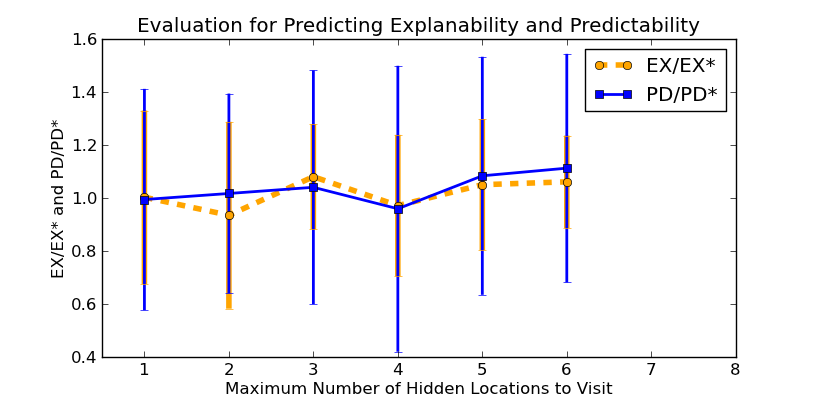}
}
\caption{Evaluation for predicting $\theta$ and $\beta$ measures as the difference
between $M_R$ and $\mathcal{M}_R^*$ increases (i.e., as the maximum number of hidden goals increases).}
\label{fig:noise}
\end{figure}

{\it Using CRFs for Plan Explicability and Predictability:}
In this evaluation,  we randomly generate $1-3$ hidden goals to include in $G'$ in $1000$ training samples.
%The number of training samples is gradually increased from $1000$ to $1900$ with step size $100$.
%After the model is trained, we evaluate it on $100$ testing samples.
%The result is presented in Fig. \ref{fig:sample}.
%We can see that the prediction performance 
%(i.e., the ratios between $\theta$ and $\beta$ computed based on $\mathcal{L}_{CRF}^*$ and $\mathcal{L}^*$) 
%is generally between $50\%-150\%$,
%which shows that our approach can capture $\mathcal{L}^*$ relatively well.  
%We anticipate this performance to be further improved when using general types of CRFs and incorporating 
%more complex features.
%Furthermore, the performance does not improve much as the sample size increases.
%This shows that our approach can capture $\mathcal{L}^*$ using a relatively small amount of training samples. 
%The bump at $1300$ in Fig. \ref{fig:sample} is due to samples with low $\beta$ values which are rare in our evaluation. 
%The number of training samples in this evaluation is $1000$.
After the model is learned, we evaluate it on $100$ testing samples
%The other settings are kept the same as those in the previous evaluation
in which we vary the maximum number of hidden goals from $1$ to $6$ with step size $1$.
The result is presented in Fig. \ref{fig:noise}.
We can see that the prediction performance 
(i.e., the ratios between $\theta$ and $\beta$ computed based on $\mathcal{L}_{CRF}^*$ and $\mathcal{L}^*$) 
is generally between $50\%-150\%$,
We can also see that the oscillation level does not seem to influence the prediction performance much.
This shows that our approach is effective 
whether $M_R$ and $\mathcal{M}_R^*$
are similar or largely different.

%\begin{figure}
%\centering
%{
%    \includegraphics{pdf/sample}
%}
%\caption{Evaluation for predicting $\theta$ and $\beta$ as the number of training samples increases.
%EX denotes $\theta$; PD denotes $\beta$. 
%The superscript $*$  is used to denote $\theta$ and $\beta$ computed from the ground truth ($\mathcal{L}^*$).
%The result shows the means and standard deviations at each setting for $100$ testing samples.}
%\label{fig:sample}
%\end{figure}

{\it Selecting Explicable and Predictable Plans:}
We evaluate plan selection using $\theta$ and $\beta$ measures 
and compare the selected plans (denoted by EXPD-SELECT) 
with plans selected by a baseline approach (denoted by RAND-SELECT). 
Given a set of candidate plans,
EXPD-SELECT selects a plan according to the highest predicted explicability or predictability measure
while RAND-SELECT randomly selects a plan from the set of candidate plans. 
%we compare the performance of planning with the consideration of $\theta$ and $\beta$ (denoted as EXPD)
%with a baseline approach that does not (denoted as BASELINE). 
To implement this, for a given public goal $G$, we randomly construct $20$ problems with a given level of oscillation 
as determined by the maximum number of hidden goals. 
Each such problem corresponds to a different $G'$ and a plan is created for it.
The set of plans for these $20$ problems associated with the same $G$ is the set of candidate plans for $G$.
For each level of oscillation, we randomly generate $50$ different $Gs$ and then construct the set of candidate plans for each $G$. 
The model here is trained with $1900$ samples using the same settings as in our first evaluation
and we gradually increase the level of oscillation. 
%The model used in this evaluation is trained using $1900$ samples using the same settings in the first evaluation. 
%While EXPD-SELECT chooses plans with the highest $\theta$ or $\beta$ values based on a model 
%learned from $1900$ samples using the same settings for generating training examples the first evaluation,
%RAND-SELECT randomly chooses plans.

\begin{figure}
\centering
{
    \includegraphics{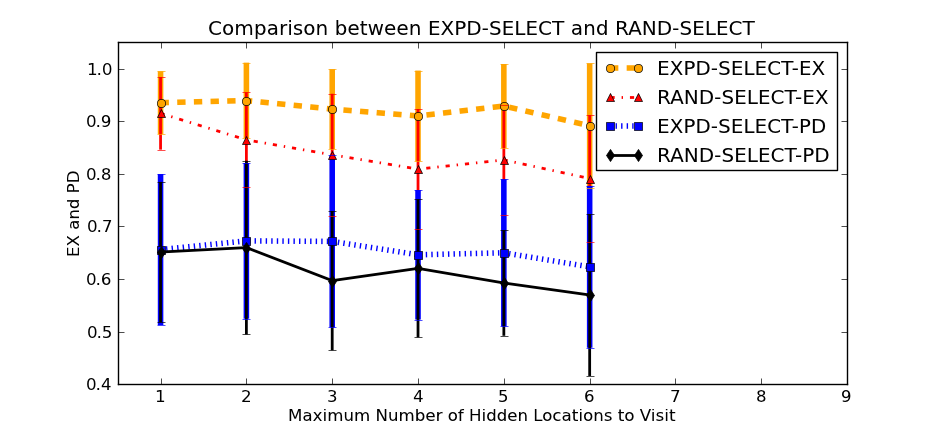}
}
\caption{Comparison of EXPD-SELECT and RAND-SELECT
}
\label{fig:plan}
\end{figure}

We compare the  $\theta$ and $\beta$ values computed from the ground truth labeling
of the chosen plans.  
The result is provided in Fig. \ref{fig:plan}. 
When the oscillation is small,
the performances of both approaches are similar.
As the oscillation increases, the performances of the two approaches diverge.
This is expected since RAND-SELECT randomly chooses plans and 
hence its performance should decrease as the oscillation increases. 
On the other hand, EXPD-SELECT is not influenced as much
although its performance also tends to decrease.
This is partly due to the fact that the model used in this evaluation is trained with samples having a maximum of $3$ hidden goals. 

In Fig. \ref{fig:plan} for explicability, almost all results are significantly different at $0.001$ level 
(except at $1$);
for predictability, results are significantly different at $0.01$ level at $3$, $5$ and $6$. 
The trend to diverge is clearly present.
Note that we use linear-chain CRFs in our evaluations, 
which does not directly model correlations among observations across states. 
These features are common in our rover domain (e.g., navigating back and forth). 
Hence, we can anticipate performance improvement with more general CRFs. 

{\it Synthesizing Explicable and Predictable Plans:}
%We have shown previously that the learned CRF can approximate the true $\theta$ and $\beta$ values for a plan. 
We evaluate here plan synthesis using Alg. \ref{alg}.
More specifically, we compare FF planner that considers the predicted $\theta$ and $\beta$ values in its heuristics
with a normal FF planner that only considers the action cost. 
The FF planner with the new heuristic is called FF-EXPD. 
In this evaluation, 
we set the maximum number of hidden locations to visit to be $6$.
For each trial, we generate $100$ problems and apply
both FF and FF-EXPD to solve the problems. 
Given that we are interested in comparing the cases when explicability is low, 
we only consider problems when the predicted plan explicability for the plan generated by FF is below $0.85$.

\begin{figure}
\centering
{
    \includegraphics{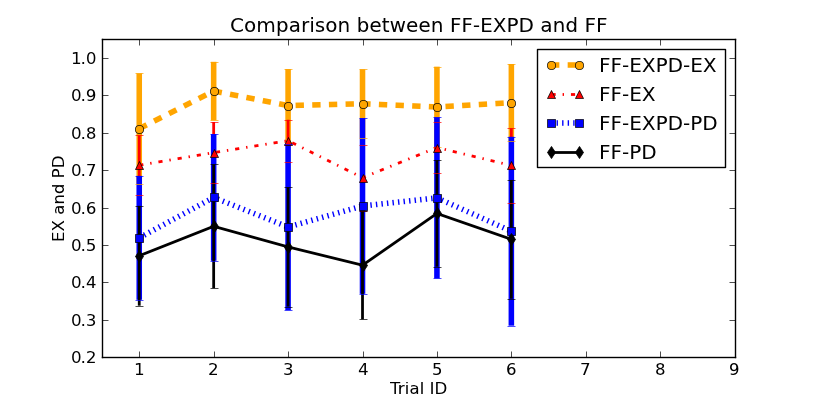}
}
\caption{Comparison of FF and FF-EXPD considering only $\theta$. 
}
\label{fig:heuristic-ex}
\end{figure}

First, we consider the incorporation of $\theta$ only. %as shown in Alg. \ref{alg}.
The result is presented in Fig. \ref{fig:heuristic-ex}.
%Given that there is a single problem in this case, 
%the variance of $\theta$ value is relatively small compared to our last evaluation. 
For the explicability measure, 
we see a significant difference in all trials.  
Another observation is that the difference in plan predictability
is present but not as significant. 
This evaluation suggests that our heuristic search can produce plans of high explicability. 

\begin{figure}
\centering
{
    \includegraphics{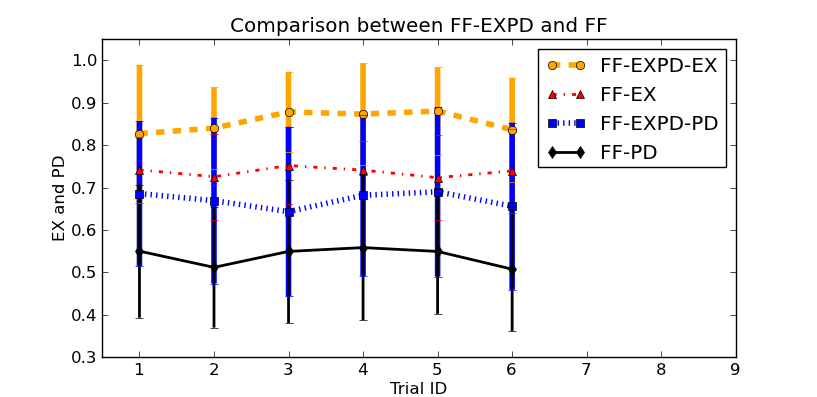}
}
\caption{Comparison of FF and FF-EXPD considering only $\beta$.
}
\label{fig:heuristic-pd}
\end{figure}

Next, we consider the incorporation of $\beta$ only.
The result is presented in Fig. \ref{fig:heuristic-pd}.
Similarly, we see a significant difference in all trials for both explicability and predictability. 
One observation is that improving on plan predictability
also improves plan explicability
which is expected given Eqs. \eqref{def:eva-ex} and \eqref{def:eva-pd}).

{\it Plan Cost:}
We consider plan cost here for the evaluation in Fig. \ref{fig:heuristic-pd}. 
The result is presented in Table \ref{tab:cost}.
%Again, we run $100$ problems for each number of hidden locations. 
%As the maximum number of hidden locations increases,
%the number of plan steps also increases for both approaches.
We can see that the plan length
for FF-EXPD is longer than the plan produced by FF in general.
This is expected since FF only considers plan cost.
However, in all settings, FF-EXPD penalizes the plan cost slightly (about $10\%$)
to improve the plan explicability and predictability measures.

\begin{table}[h!]
  \begin{center}
    \caption{Plan Steps Comparison for Fig. \ref{fig:heuristic-pd}}
    \label{tab:cost}
    \begin{tabular}{|l|c|c|c|c|c|c|}
    \hline
      Trial ID & 1 & 2 & 3 & 4 & 5 & 6\\
      \hline
      FF (avg. \# steps) & 21.9 & 24.0 & 24.1 & 23.9 & 22.1 & 22.4\\
      FF-EXPD (avg. \# steps) & 23.5 & 26.3 & 25.2 & 24.0 & 23.4 &25.0\\
      \hline
    \end{tabular}
  \end{center}
\end{table}

% !TEX root = ../explanation.tex

\subsection{Evaluation with Physical Robots}
\label{car}

In this section we evaluate our approach in a blocks world domain with a physical robot. 
It simulates a smart manufacturing environment where robots are working beside humans. 
Although the human and robot do not have direct interactions -- the robot's goal is independent of the human's,
generating explicable plan is still an important issue since it will help humans concentrate more on their own tasks. 
%where the concept of plan explicability plays a significant role during plan execution -- driving, where the failure to be explicable to other drivers in the surrounding cars has major consequences on safety.
%and hence much of such explicable behaviors has, in fact, been made mandatory by law. 
%This becomes particularly relevant for self-driving cars which must learn to adapt to human behaviors so as to ensure that it is not always performing what is optimal for itself but also making its actions intelligible to humans around it.
%For example, to ensure its own safety, the agent may produce an oscillating behavior (to avoid collision based only on its own model and sensor information) when it needs to change lanes and there is another car tailing closely.
%In this evaluation, we will attempt to learn patterns from real data generated by humans on preferred (and consequently safer for all agents) ways to change lanes such that the actions are explicable to other cars (driven by humans).
Here, we evaluate plans generated by the robot using FF-EXPD and a cost-optimal planner (OPT) in various scenarios 
and compare the plans with human subjects in terms of explicability.

\subsubsection{Domain Description}

In this domain, the robot's goal (which is known to the human) is to build a tower of a certain height using blocks on the table.
The towers to be built have different heights in different problems. 
There are two types of blocks, light ones and heavy ones,
which are indistinguishable externally but the robot can identify them based on the markers. 
Picking up the heavy blocks are more costly than the light blocks for the robot. 
Hence, the robot may sometimes choose seemingly more costly (i.e., longer) plans
to build a tower from the human's perspective. 

\subsubsection{Experimental Setup}

%The experiment setup is shown in Fig. \ref{}. 
We generated a set of $23$ problems in this domain in which towers of height $3$ are to be built. 
The plans for these problems were manually generated and labeled as the training set.
For $4$ out of these $23$ problems, the optimal plan is not the most explicable plan. 
To remove the influence of grounding,
we also generated permutations of each plan using different object names for these $23$ problems,
which resulted in a total of about $15000$ training samples.
We then generated a set of $8$ testing problems for building towers of various heights (from $3-5$) 
to verify that our approach can generalize.  
Testing problems were generated only for cases where plans are more likely to be inexplicable. 
For each problem, we generated two plans, 
one using OPT and the other using FF-EXPD,
and recorded the execution of these plans on the robot.
We recruited $13$ subjects on campus and
each human subject was tasked with labeling two plans (generated by OPT and FF-EXPD respectively) for each of the $8$ testing problems,
using the recorded videos and following a process similar to that used in preparing training samples.
After labeling each plan, we also asked the subject to provide a score ($1-10$ with $10$ being the most explicable) 
to describe how comprehensible the plan was overall. 

%This domain will contain scenarios in which the optimal plan would be 
%
%For the experiments we use a 2D simulator such as the one shown in Fig. \ref{car}. The simulator has two cars driving in adjacent lanes. 
%The human interacting with the interface observes from the perspective of the blue car, 
%while the red car (i.e., the autonomous agent) executes a sequence of actions to change into the same lane as the human. 
%The red car can move sideways at different granularities (i.e., move left and right a bit at a time or changing lane in one step), and turn on or off one or more of its rear lights. 
%
%We randomly generated $130$ plans with various lengths and combinations of actions to achieve the lane changing goal. 
%The human subject is informed that the red car is to change lane
%and is tasked with labeling the actions of the autonomous car as being part of a \emph{``changing lane"} task or as \emph{``inexplicable"}. 
%For each of these plans, we asked a human subject to provide the label information while having the red car executing the plan in the simulator.  
%Following are the results of the study given the annotated plan traces from $13$ human subjects. 

\subsubsection{Results}

\begin{figure}
\centering
{
    \includegraphics{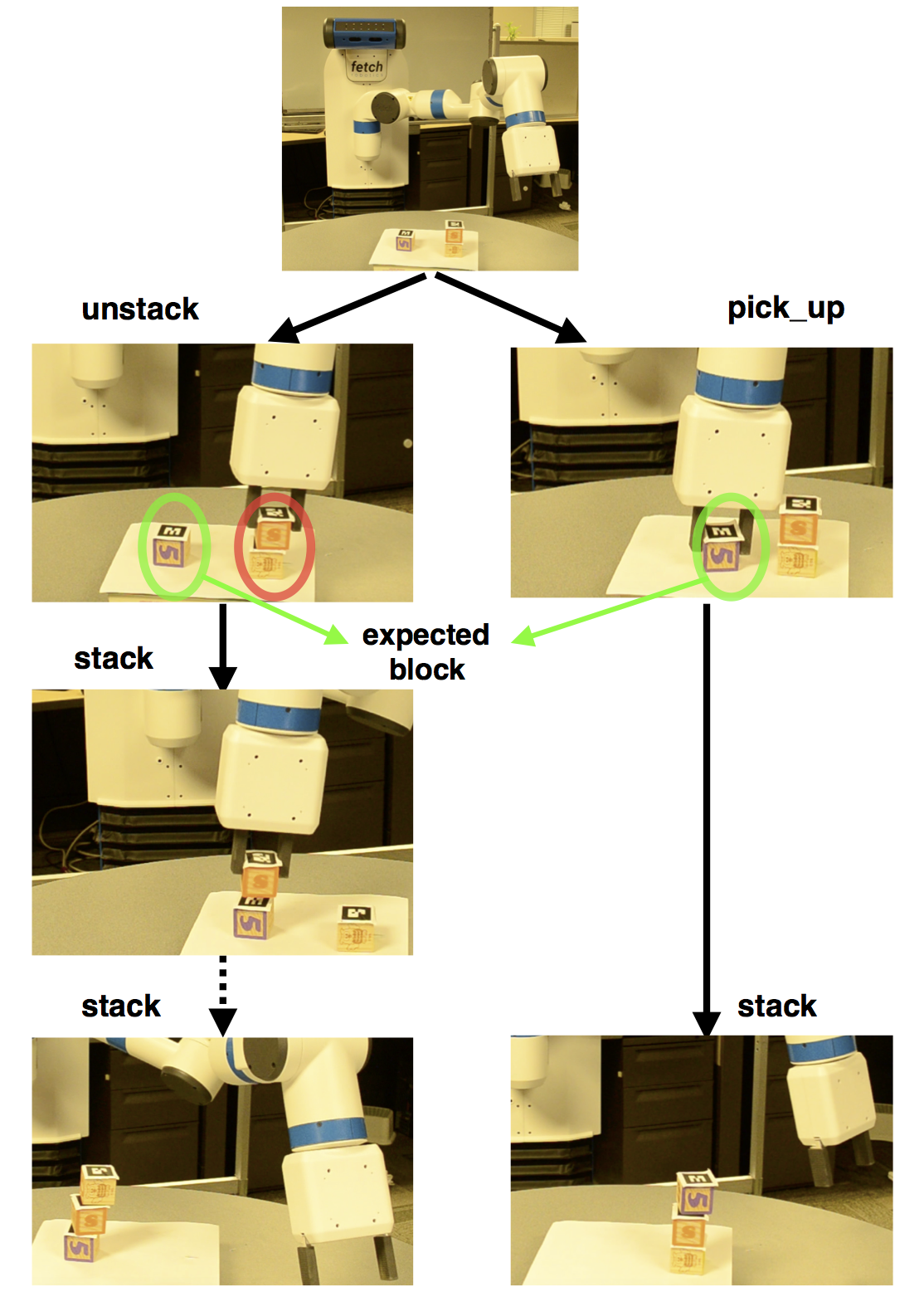}
}
\caption{Plan execution of two plans generated by OPT (left) and FF-EXPD (right) for one out of the $8$ testing scenarios.
The top figure shows the setup of this scenario where the goal is to build a tower of height $3$.
The block that is initially on the left side of the table is a heavy block.
The optimal plan involves more actions with the light blocks 
(i.e., putting the two light blocks on top of the heavy one)
while the explicable plan is more costly since it requires moving the heavy one. }
\label{fig:robot}
\end{figure}

In this evaluation, we only use one task label ``$building$ $tower$''.
For all testing problems, the labeling process results in $77.8\%$ explicable actions (i.e., actions with a task label) 
for OPT and $97.3\%$ explicable actions for FF-EXPD. 
The average explicability measures for FF-EXPD and OPT are $0.98$ and $0.78$, 
and the average scores are $9.65$ and $6.92$, respectively.
We analyze the results using a paired T-test which shows a significant difference between FF-EXPD and OPT
in terms of the explicability measures (using Eq. \eqref{def:eva-ex}) computed from the human labels and the overall scores  
($p < 0.001$ for both). 
Furthermore, after normalizing the scores from the human subjects,
the Cronbach's $\alpha$ value shows that the explicability measures and the scores are consistent for both FF-EXPD and OPT
($\alpha = 0.78, 0.67$, respectively).
%no significant difference is found between the explicability measures and the scores for either FF-EXPD or OPT ($p = 0.137, b = $, respectively). 
These results verify that: $1)$ our explicability measure does capture the human's interpretation of the robot plans
and $2)$ our approach can generate plans that are more explicable to humans. 
In Fig. \ref{fig:robot}, we present the plans for a testing scenario.
The left part of the figure shows the plan generated by OPT
and the right part shows the plan generated by FF-EXPD.
A video is also attached showing the different behaviors with the two planners
in this scenario. 

\section{Conclusion}

While we are still far from having intelligent robots and agents working side-by-side of humans as teammates (rather than as tools), 
it becomes increasingly important to consider issues when such autonomous agents appear in our everyday life. 
These agents need to create and execute complex plans.
In this paper, we introduced plan explicability and predictability for such agents
so that they can synthesize plans that are more comprehensible to humans.
To achieve this, 
they must consider not only their own models 
but also the human's interpretation of their models.
%This enables agents to synthesize plans that can be easily understood by humans.
To the best of our knowledge, this is the first attempt to 
model plan explicability and predictability for task planning
which differs from previous work on human-aware planning.
The proposed measures have a variety of applications
(e.g., achieving fluent human-robot interaction and ensuring human safety).
To compute these measures, 
we learn the labeling scheme of humans for agent plans from training examples based on CRFs.
We then use this learned model to label a new plan to compute its explicability and predictability.
The proposed approach is evaluated on a synthetic domain and with human subjects using physical robots
to show its effectiveness.
%to show that it is effective in capturing explicability and predictability,
%and producing explicable and predictable plans.
%Our evaluations also demonstrate the generality of our approach in modeling various domains. 
A natural extension of our work is to consider human-robot teaming
where there exists close interactions.
Humans in our current settings are observers. 

Finally, while we focus on the explicability and predictability measures for robot task planning, 
they also have many other interesting applications. 
For example, many defense applications use planning to create unpredictable and inexplicable plans, 
which can help deter or confuse enemies 
and are also useful for testing defenses against novel or unexpected attacks.
These applications can be implemented using our approach 
by minimizing the $\theta$ and $\beta$ measures
instead of maximizing them.

%\section*{Acknowledgments}

%% Use plainnat to work nicely with natbib. 

\bibliographystyle{plainnat}
\bibliography{explanation}

\end{document}